\documentclass{article}
\usepackage{spconf,amsmath,graphicx,booktabs}
\usepackage{multirow}
\usepackage{bbm}
\usepackage{url}
\usepackage[utf8]{inputenc}
\usepackage{xcolor}
\usepackage{blindtext}
\usepackage{etoolbox}
\makeatletter
\patchcmd{\@verbatim}
  {\verbatim@font}
  {\verbatim@font\small}
  {}{}
\makeatother

\usepackage[belowskip=-10pt,aboveskip=6pt]{caption}
\def\LL{Libri-light}

\title{Libri-Light: A Benchmark for ASR with Limited or No Supervision}

\name{ J. Kahn$^{*\dagger{}}$, M. Rivière$^{*\dagger{}}$, W. Zheng$^{*\dagger{}}$, E. Kharitonov$^{*\dagger{}}$, Q. Xu$^{*\dagger{}}$, P.E. Mazaré$^{*\dagger{}}$,  J. Karadayi$^{*\ddagger{}}$}{\it V. Liptchinsky$^{\dagger{}}$, R. Collobert${^\dagger{}}$, C. Fuegen$^{\dagger{}}$, T. Likhomanenko$^{\dagger{}}$, G. Synnaeve$^{\dagger{}}$, A. Joulin$^{\dagger{}}$,}{\it  A. Mohamed${^\dagger{}}$, E. Dupoux$^{\dagger{}\ddagger{}}$\thanks{* Contributed equally, in random order.}}

\address{$^\dagger{}$Facebook AI, $^\ddagger{}$EHESS, ENS, PSL-University, CNRS, INRIA}
\begin{document}

\maketitle

\begin{abstract}
We introduce a new collection of spoken English audio suitable for training speech recognition systems under limited or no supervision. It is derived from open-source audio books from the LibriVox project. It contains over 60K hours of audio, which is, to our knowledge, the largest freely-available corpus of speech. The audio has been segmented using voice activity detection and is tagged with SNR, speaker ID and genre descriptions. Additionally, we provide baseline systems and evaluation metrics working under three settings: (1) the zero resource/unsupervised setting (ABX), (2) the semi-supervised setting (PER, CER) and (3) the distant supervision setting (WER). Settings (2) and (3) use limited textual resources (10 minutes to 10 hours) aligned with the speech. Setting (3) uses large amounts of unaligned text. They are evaluated on the standard LibriSpeech dev and test sets for comparison with the supervised state-of-the-art.  
\end{abstract}


\section{Introduction}
\label{sec:intro}
\vspace{-2mm}
Automatic Speech Recognition (ASR) has made striking progress in the recent years with the deployment of increasingly large deep neural networks trained on increasingly large sets of annotated speech (from thousands to tens of thousands of hours). This approach is hit by diminishing returns as the costs of annotating even larger datasets become  prohibitive. It is also difficult to scale beyond a handful of high-resource languages and address the needs of a long tail of low-resource languages, dialectal and idiolectal variants, accents, and registers. As such, there has been a recent surge of interest in weakly supervised solutions that use datasets with fewer human annotations. In the semi-supervised setting, only a fraction of the dataset is labelled and the rest is unlabelled \cite{tur2005,kahn2019self}, while in a distant supervision setting, the dataset is mostly or entirely unlabelled, but large quantities of unaligned text provide a language model corpus \cite{chen2018almost,chung2018}. Other approaches have addressed pretraining with labels from other languages \cite{
huang2013cross,vesely2012language} or pretraining using unsupervised objectives \cite{vandenoord2018,schneider2019wav2vec}. At the extreme of this continuum, zero resource ASR discovers its own units from raw speech \cite{Versteegh2016,dunbar2017,dunbar2019}. Despite many interesting results, 
the field lacks a common benchmark (datasets, evaluations, or baselines) for comparing ideas and results across these settings. Here, we introduce \LL{}, a large open-source corpus (60K hours) of unlabelled speech and a common set of metrics to evaluate three settings: (1) the zero-resource/unsupervised setting (ABX), (2) the semi-supervised setting (PER and CER), and (3) the distant supervision setting (WER). The last two settings use a limited-resource training set (10 min, 1h, 10h), and the last one large in-domain and out-of-domain text to train language models. The test sets are identical to LibriSpeech \cite{librispeech2015} so as to facilitate comparison of weakly supervised results with the state-of-the art in supervised learning. We also provide a baseline system on these three settings. All datasets, metrics and baseline systems are open source\footnote{\url{https://github.com/facebookresearch/libri-light}}.

\vspace{-3mm}
\section{Related work}\label{sec:related}
\vspace{-3mm}
The release of open source software and datasets has facilitated rapid progress in machine learning and in particular large vocabulary ASR. LibriSpeech is one of the first large-scale open-source datasets and contains over 1000 hours of audio books, together with textual annotations aligned at the sentence level. Mozilla's CommonVoice project has facilitated data collection across several languages and currently contains 2900 hours of read speech in 37 languages\footnote{\url{https://voice.mozilla.org}}. A. Black at CMU has compiled the Wilderness dataset which consists of the text of the Bible read in 750 languages \cite{black2019cmu}. Other open-source resources are available from OpenSLR\footnote{\url{http://openslr.org/}}.

The Zero Resource Challenge has released a series of datasets and metrics for the unsupervised setting \cite{Versteegh2016,dunbar2017}\footnote{\url{https://zerospeech.com}}, but the datasets are generally small (between 2.5 and 50~h). In this work, we substantially expand dataset size and use the same evaluation metrics (ABX \cite{Schatz2013ABX}) for comparability. The IARPA Babel program \cite{harper2014} has also initiated a push towards limited supervision for less studied languages. In its most difficult track, the dataset contains only 10 hours of transcribed speech in conjunction with  with larger amounts of untranscribed audio. Here, we retain 10 hours as a \textit{upper} bound, and add lower-resource sets containing 1 hours and 10 minutes of labeled audio. While distant supervision has been the focus of two JHU-JSALT workshops (2016 \cite{liu2017empirical}, 2019 \cite{chorowski2019}) but no benchmark has yet emerged. 

\vspace{-5pt}
\section{Dataset and metrics}
\label{sec:dataset}
\vspace{-5pt}
\subsection{Dataset}
\vspace{-10pt}

\begin{figure}[h]
\begin{minipage}[b]{1.0\linewidth}
  \centering
  \centerline{\includegraphics[width=6.0cm]{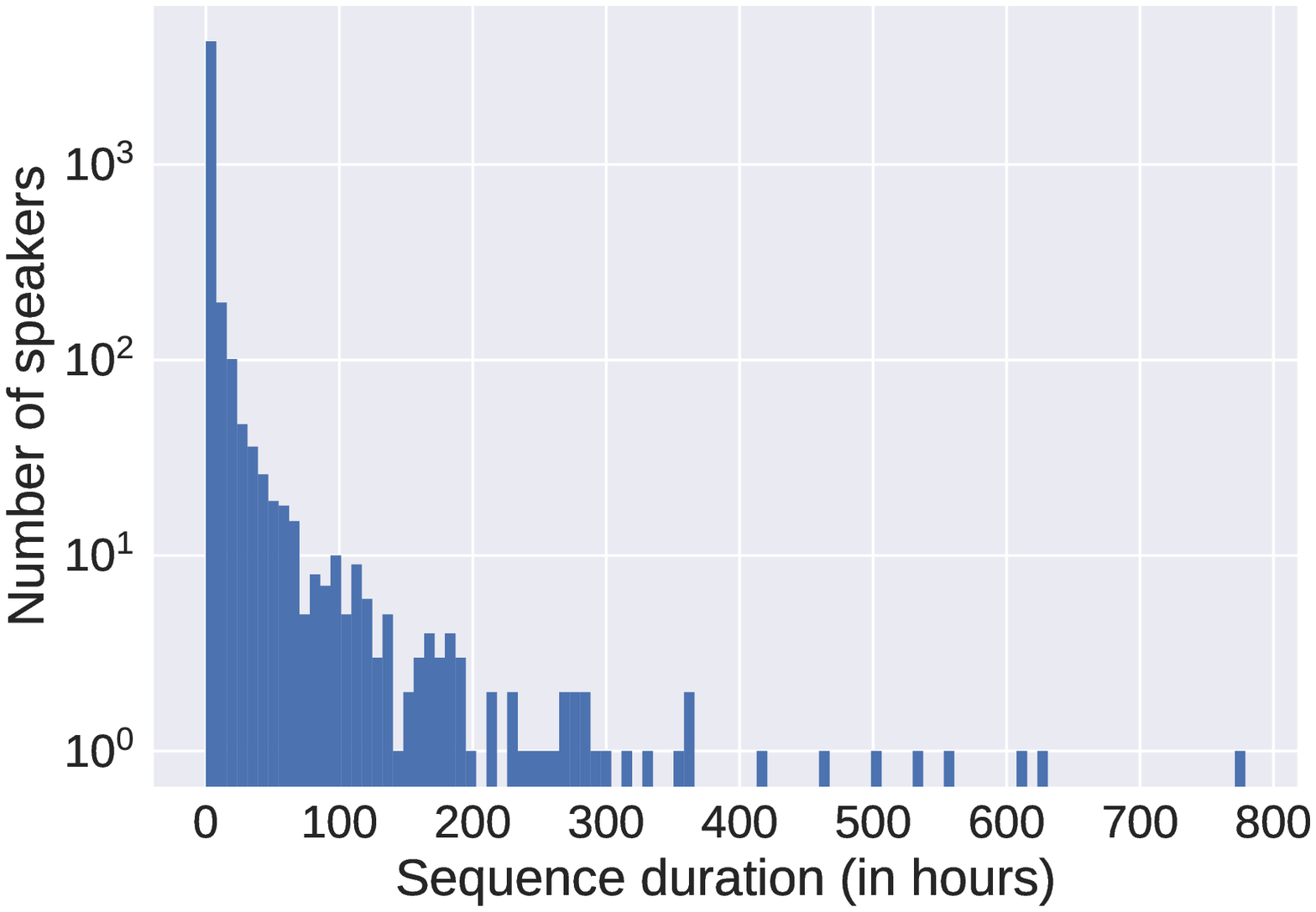}}
  \vspace{-1.3mm}
  \centerline{(a)}\medskip
\end{minipage}
\begin{minipage}[b]{1.0\linewidth}
  \centering
  {\includegraphics[width=6.0cm]{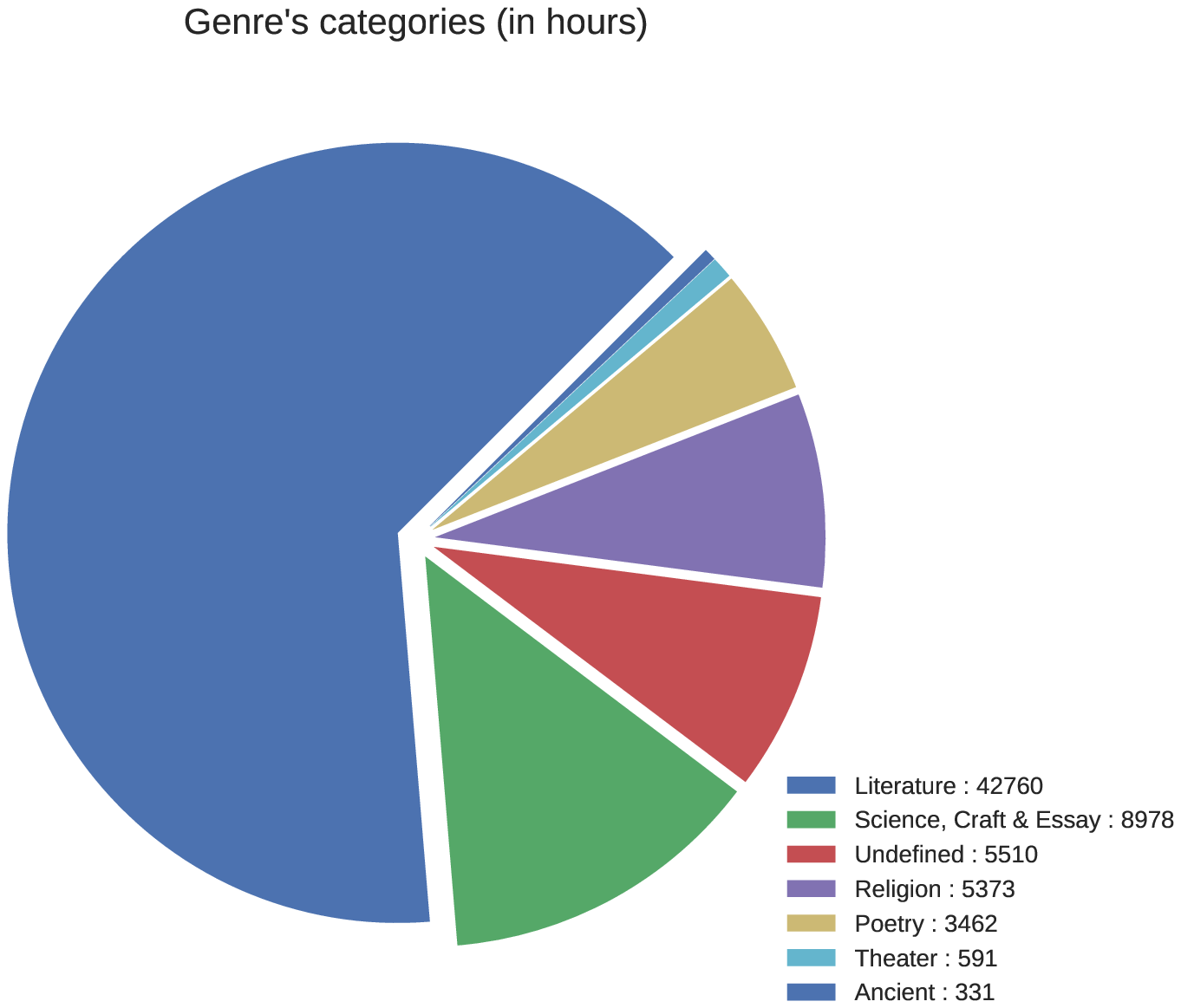}}
  \centerline{(b)}\medskip
\end{minipage}
\caption{\textbf{Corpus statistics.} (a) Durations in hours per speakers (b) Durations for the 25 most frequent genres.}
\label{fig:corpus}
\end{figure}

\begin{table}[t]
\centering
\begin{tabular}{l @{\hspace{0.8\tabcolsep}} c @{\hspace{0.9\tabcolsep}} c @{\hspace{0.9\tabcolsep}} c @{\hspace{0.9\tabcolsep}} c @{\hspace{0.9\tabcolsep}} c}
\toprule
subset          & hours & books & files & per-spk & total \\
                &       &       &     & hours& spkrs \\
\midrule
    \multicolumn{6}{l}{\textit{Unlabelled Speech Training Set}}\\
unlab-60k   & 57706.4 &   9860 & 219041   & 7.84 &  7439  \\
unlab-6k    & 5770.7  &   1106  & 21327   & 3.31 &  1742 \\
unlab-600   & 577.2   &   202   & 2588    & 1.18 &  489  \\
\midrule
\toprule
subset          & hours & per-spk & female& male & total \\
                &       & minutes & sprks & spkrs& spkrs \\
\midrule
\multicolumn{6}{l}{\textit{Limited Resource Training Set}}\\
train-10h       &10   & 25  &  12  &    12  &   24    \\
train-1h        &1    & 2.5 &  12  &    12  &     24  \\
train-10m*      &10min& 2.5 &  2   &    2   &    4   \\
\midrule
\multicolumn{6}{l}{\textit{Dev \& Test Sets (from LibriSpeech)}}\\
dev-clean       & 5.4 &   8  &  20  &  20 & 40 \\
dev-other       & 5.3 &   10 &  16  &  17 & 33 \\
test-clean      & 5.4 &   8  &  20  &  20 & 40 \\
test-other      & 5.1 &   10 &  17  &  16 & 33 \\
\midrule
\midrule
\multicolumn{3}{l}{subset}          & tokens &vocab  & \\
\midrule
    \multicolumn{6}{l}{\textit{Unaligned Text Training Set}}\\
\multicolumn{3}{l}{librispeech-LM (in-domain)} & 800M      & 200K  &    \\
\bottomrule
\end{tabular}
\caption{\textbf{Datasets stats in \LL{}}. *Six different versions of the 10 min datasets have been constructed, the union of these small datasets make up the 1h dataset.}\label{tab:corpus}
\end{table}

The dataset is composed four parts: a train set with unlabelled speech, a train set with limited labels, dev/test sets, and a train set containing unaligned text; see  Table~\ref{tab:corpus}. 

\begin{table*}[]
\begin{center}
\begin{tabular}{l c @{\hspace{0.8\tabcolsep}} c @{\hspace{0.8\tabcolsep}} c @{\hspace{0.8\tabcolsep}} cc  c  @{\hspace{0.8\tabcolsep}}c @{\hspace{0.8\tabcolsep}} c @{\hspace{0.8\tabcolsep}} c }
\toprule
  & \multicolumn{4}{c}{{ABX within speaker}}  && \multicolumn{4}{c}{{ABX across speaker}} \\
\cline{2-5}\cline{7-10}
System               &dev-clean&dev-other&test-clean&test-other&&dev-clean&dev-other&test-clean&test-other\\
\midrule
MFCC Baseline   & 10.95   &  13.55  & 10.58    & 13.60    &&  20.94  & 29.41   & 20.45 &  28.5  \\
CPC unlab-600   &  7.36   & 9.39    &  6.90    &   9.59   &&    9.58 & 14.67    & 9.00 &  15.1    \\ 
CPC unlab-6k    &   6.51 &   8.42   &  6.22    &  8.55    &&  8.48   &  13.39  &8.05    & 13.81  \\
CPC unlab-60k   &\bf6.11 &\bf 8.17  &\bf 5.83  &\bf 8.14  &&\bf 8.05 &\bf 12.83&\bf 7.56&\bf 13.42 \\ 
\bottomrule
\end{tabular}
\caption{\textbf{ABX errors on unsupervised CPC trained features.} Within- and across-speaker phoneme discriminability scores (lower is better) on the LibriSpeech dev and test sets as a function of varying quantities of unlabelled speech.}\label{tab:zero}
\end{center}\vspace{-10pt}

\end{table*}

\textbf{Unlabelled Speech Training Set}. This dataset was obtained by extracting audio files for English speech from the LibriVox repository\footnote{\url{https://librivox.org}} containing open source audio books.  Files were downloaded and converted to 16kHz FLAC. We then removed corrupted files, files with unknown or multiple speakers, and speakers appearing in LibriSpeech dev and test sets. The potentially duplicated versions of books based on titles were set aside (and distributed as a \texttt{duplicate} subset, totalling 4500 hours). We then ran a Voice Activity Detection (VAD) model using the wav2letter++ framework~\cite{wav2letter++} on the recordings to tag onsets and offsets of speech segments
. The VAD segments were used to derive an average SNR for each file. For each file, we constructed JSON metadata including title, unique speaker ID, SNR, genre, and list of valid VAD\_block (block of more than 500ms of speech indicated by onsets and offsets). We created three dataset splits based on different sizes: (\texttt{unlab-60k}), (\texttt{unlab-6k}) and (\texttt{unlab-600}), matched in genre distribution (the smaller cuts are included in the larger ones, see the stats in Table~1).
The distributions by speaker and genres are in Figure~\ref{fig:corpus}. The total amount of speech in the dataset is 62.2K hours, including the \texttt{duplicate} files.
 
\textbf{Limited-resource Training Set}. For training with limited supervision, we selected three subsets of the LibriSpeech training set: a 10 hour set, a 1 hour set, and six 10-minute sets (the six 10-minute sets together make up the 1h set, and the 1h set is included in the 10h set). In each set, half of the utterances are from the clean and other training sets, respectively. We additionally provide orthographic transcriptions from LibriSpeech and phonetic transcriptions generated from phonemizer\footnote{\url{https://gitlab.coml.lscp.ens.fr/mbernard/phonemizer}}. 

\textbf{Dev and Test Set}. The dev and test sets are the same as that of LibriSpeech (5.4 hours for dev-clean, 5.3 hours for dev-other, 5.4 hours for test-clean, and 5.1 hours for test-other) and are intended for testing and tuning. All dev or test set audio has been removed from training sets. The ground-truth phonetic sequences for the dev and test sets were generated as above; in addition, for ABX evaluation, forced alignment was obtained using a model trained on LibriSpeech.

\textbf{Unaligned Text Training Set}. For training a language model in the distant supervision setting, we consider the LM corpus provided in LibriSpeech\footnote{\url{https://openslr.org/11/}} which contains 800M tokens and a vocabulary size of 200k from 14.5k public books from Project Gutenberg
. This corpus only partially overlaps with the content of our unlabelled training set and can thus be considered in-domain. Several options exist for publicly available out-of-domain corpora (wikitext103, 1Billion word, etc). 

\vspace{-8pt}
\subsection{Metrics}
\label{sec:metrics}
\vspace{-5pt}

We provide 3 sets of metrics for the unsupervised, distantly-supervised, and semi-supervised settings. 

For the unsupervised setting, the aim is to extract speech representations (discrete or continuous) which encode the phonetic content of the language while ignoring irrelevant information (channel, speaker, etc). The representation is evaluated using  ABX error, a distance-based metric used in previous zero resource challenges \cite{Versteegh2016,dunbar2017,dunbar2019}. For a given pair of sounds (for instance, "bit" versus "bet"), we compute the probability that the distance between a random token of "bit" (X) is closer to another token of "bit" (A) than to a token of "bet" (B). The ABX error rate is obtained by averaging across all such minimal pairs of phone trigrams in the corpus. For the ``within-speaker'' score, A, B and X are from the same speakers; in the ``across-speaker'' score, A and B are from the same speaker, but X is from a different speaker (see \cite{schatz2016abx}).

For the semi-supervised setting, we evaluate the quality of learned acoustic representations with little annotated data. Models can be trained either with character or phonetic targets using limited data and measured by either Character Error Rate (CER) or Phoneme Error Rate (PER).

For distant supervision, we evaluate how the learned representations can be used to decode speech at the word level using a pre-trained language model. We use Word Error Rate (WER) for the evaluation. Because the dev and test sets are from LibriSpeech, this allows to compare distant supervision directly with SOTA supervised models. More details on dataset and metrics in Supplementary Section~\ref{sec:supmeth}. 

\begin{table}[h]
\begin{tabular}{l c @{\hspace{0.8\tabcolsep}} c @{\hspace{0.8\tabcolsep}} c @{\hspace{0.8\tabcolsep}} c }
\toprule
                          & dev-      & dev-    & test-     & test-\\
System                    & clean     & other   & clean     & other\\
\midrule
no pretraining+train-10h  &  45.9     &   55.7  &   43.7    &    58.6  \\
CPC unlab-60k+train-10m   &  40.1     &   51.5  &   39.4    &    53.3  \\
CPC unlab-60k+train-1h    &  32.2     &   44.6  &   31.6    &    46.8  \\
CPC unlab-60k+train-10h   & \bf{28.4} &\bf{41.4}& \bf{27.9} & \bf{43.6}\\
\bottomrule
\end{tabular}
\caption{\textbf{PER/CER in the semi-supervised setting.} A pretrained CPC system plus a linear classifier trained on limited amounts of labels compared to the same system trained from scratch (PER). 
} 
\label{tab:semi}
\end{table}
\begin{table}[h]
\begin{tabular}{l @{\hspace{0.6\tabcolsep}} c @{\hspace{0.6\tabcolsep}} c @{\hspace{0.8\tabcolsep}} c @{\hspace{0.6\tabcolsep}} c}
\toprule
                      &dev-  &dev- &test-  &test-\\
System                & clean&other&clean&other\\
\midrule
\midrule
\multicolumn{5}{l}{\textit{Supervised systems (LibriSpeech 1000 h)}} \\
Gated Cnv+4gramLM\cite{liptchinsky2017based}    &  4.6  &  13.8 &  4.8  &  14.5    \\
Hybrid+seqdisc+4gramLM\cite{luscher2019rwth}   &   3.4  & 8.3   &   3.8 &   8.8    \\
\midrule
\midrule
\multicolumn{5}{l}{\textit{CPC pretrain + CTC fine-tuning + 4gram-LM}}   \\
CPC unlab-600+train-10m    &     97.3  &     97.6  &     97.1  &   97.7  \\
CPC unlab-600+train-1h     &     72.2  &     84.5  &    70.1  &   86.3  \\
CPC unlab-600+train-10h    &     52.5  &     71.6  &     49.3  &   74.1  \\
\midrule
CPC unlab-6k+train-10m     &     93.6  &     95.2  &     93.2  &   94.9  \\
CPC unlab-6k+train-1h      &     67.5  &     81.3  &     65.4  &   82.0  \\
CPC unlab-6k+train-10h     &     46.4  &    \bf 66.7  &  44.7  &   69.3  \\
\midrule
CPC unlab-60k+train-10m    &     92.5  &    94.2  &     92.5  &     94.4 \\
CPC unlab-60k+train-1h     &     66.6  &     80.0  &     64.7  &     81.6 \\
CPC unlab-60k+train-10h    & \bf 46.1  & \bf 66.7  & \bf 43.9  & \bf 69.5 \\
\midrule
\midrule 
\multicolumn{5}{l}{\textit{MFSC + TDS + CTC + Grapheme + 4gram-LM}} \\
train-1h               &    79.4     &     88.1    &      78.4    &      88.0   \\
~~ + 60k pseudo-label  &    78.6     &     86.5    &      77.2    &      86.3   \\
train-10h              &    34.0     &     60.9    &      33.5    &      62.1   \\
~~ + 60k pseudo-label  &    30.5     &     55.8    &      30.1    &      57.2   \\
\midrule
\multicolumn{5}{l}{\textit{MFSC + TDS + CTC + Phoneme + 4gram-LM}} \\
train-1h               &     81.1    &     88.5    &      80.2    &      88.7   \\
~~ + 60k pseudo-label  &     84.3    &     90.0    &      84.0    &      90.5   \\
train-10h              &     44.1    &     64.2    &      43.8    &      65.1   \\
~~ + 60k pseudo-label  &  \bf 30.0   & \bf 55.8    &  \bf 29.3    & \bf  56.6   \\
\bottomrule
\end{tabular}
\caption{\textbf{WER in the distant supervision setting.} Top: State-of-the-art supervised systems using our 4-gram-LMs. Middle: A CPC system trained with unlabelled speech, fine-tuned with limited data and integrated with a 4-gram word language model (Librispeech-LM). Bottom: A small MFSC TDS system trained on limited labeled data (graphemes or phonemes). The pseudo-labels for the 60k corpus segmented into 36-second chunks are generated and are used to retrain a larger TDS system.}\label{tab:pseudo}\label{tab:distant} 
\vspace{-5pt}
\end{table}

\vspace{-5pt}
\section{Baseline Systems}
\label{sec:pagestyle}
\vspace{-5pt}

In the unsupervised setting, we use a PyTorch implementation of the Contrastive Predictive Coding (CPC) system \cite{vandenoord2018} trained to predict the hidden states of N future speech frames and containing an encoder, a sequence model, and a predictor. The encoder maps waveforms to hidden states (one 512 dimensional embedding every 10 ms frames) using a stack of 5 convolutional layers. The sequence model encodes the hidden states into a 512-dimensional phonetic embedding with one layer of Gated Recurrent Units (GRUs). The predictor maps the last phonetic embedding onto a future hidden state using a linear projection (one linear projection per time step, varying from 1 to 12). To avoid collapsing to a trivial solution, the model is trained discriminatively; the loss function aims at decreasing the dot product between predicted and actual future frames while increasing it for frames belonging to negative sequences (distant time windows). We used a reimplementation of the original paper, which we modified to increase stability and performance, as we could not reproduce the original paper results with the provided descriptions. These changes are as follows: we replaced batch-norm with channel-wise normalization, we reduced the hidden and phonetic embeddings to 256 dimensions, used a LSTM instead of a GRU, and used a 1-layer transformer network instead of a linear projection. The original paper obtained 65.5\% accuracy on phoneme classification with a linear classifier trained on top of the frozen system's phonetic embedding. Our modified system obtained 68.9\% accuracy, while using 4 times fewer parameters in the encoder+sequence model part of the system. We trained it on the three cuts (\texttt{unlab-600}, \texttt{unlab-6k} and \texttt{unlab-60k}). 



In the semi-supervised setting, we use our baseline pretrained CPC system supplemented with a linear classifier trained with CTC loss on the limited-resource set's phone labels (only the linear layer is fine-tuned). We also provide a from-scratch control with the same architecture trained end-to-end. 

For the distant supervision setting, we run two experiments: (1) we use our pretrained CPC system with an improved CTC layer (LSTM) which we fine-tune with orthographic labels in the limited-resource set. We decode with a python wrapped version of the wav2letter++ decoder \cite{wav2letter++}, using a 4-gram KenLM \cite{heafield2011kenlm} language model trained on the unaligned text set. (2) 
We use CTC to train small Mel-filterbanks-based TDS systems\cite{hannun2019tds}, (7 TDS blocks, 20M parameters, total stride 2) on phonemes/graphemes respectively. We create pseudo-labels by beam-search decoding the 60k-hours unlabelled data with a 4-gram KenLM decoder trained on LibriSpeech-LM. These labels are used to train larger TDS systems (11 TDS blocks, 37M parameters) from scratch which generate WERs when decoding with the same LM. More details on baselines in Supplementary Section~\ref{sec:supbaseline}.

\vspace{-8pt}
\section{Results}
\label{sec:typestyle}
\vspace{-5pt}
The results for the unsupervised setting are shown in Table~\ref{tab:zero}. CPC constructs embeddings with good ABX scores compared to an MFCC baseline, and are in the same range as the SOTA in the Zero Resource Speech Challenge 2017 for English (6.2\% within and 8.7\% across \cite{heck2017}). The results in the semi-supervised setting (Table~\ref{tab:semi}) show gains in PER in using unsupervised pretraining for several different amounts of fine tuning. 
The results on the distant supervision (Table~\ref{tab:distant}), while far from supervised state-of-the-art, show that increasing the amount of unsupervised pretraining helps. Pseudo-labels are beneficial but only if the generating and fine-tuned models are initially good (Table~\ref{tab:semi} and \ref{tab:distant}).


\vspace{-8pt}
\section{Conclusion}
\label{sec:conc}
\vspace{-5pt}
We have introduced a new large dataset for benchmarking ASR systems trained with limited or no supervision. We found that unsupervised training with increasingly larger dataset yield better features and can significantly boost the performance of systems trained with limited amounts of labels (from 10 min to 10 hours) both for a phoneme recognition task in a semi-supervised setting and for a word recognition in a distant-supervision setting. The baselines were not particularly optimized for the tasks and are provided only as a proof-of-concept; there is a significant margin with fully-supervised systems. Obvious improvements include using larger models, speaker-adversarial losses, fine tuning the entire system (not just the top layers), and pseudo-labels retraining in all settings. Active learning \cite{hakkani2002active} could further select useful parts of the dataset (we have provided SNR data to facilitate this effort). Yet another approach might apply language modeling techniques directly on unlabelled audio to improve the representations before fine-tuning them \cite{chung2018speech2vec,anonymous2020vqwavvec}.

\vspace{-5pt}

\bibliographystyle{IEEEbib}
\bibliography{librilight}

\setcounter{section}{0}
\setcounter{table}{0}
\setcounter{figure}{0}

\renewcommand\thesection{S\arabic{section}}
\renewcommand\thetable{S\arabic{table}}
\renewcommand\thefigure{S\arabic{figure}}

\section{Supplementary Datasets and Metrics}\label{sec:supmeth}
\subsection{Datasets and meta-data}

The dataset is constructed according to the following pipeline:
data download, exclusion of bad files, conversion to flac, extraction of VAD, SNR, and Perplexity
, the construction of JSON files and the split in three cuts of varying sizes.  

\subsubsection{VAD}\label{sec:supvad}
Voice Activity Detection is accomplished using a TDS acoustic model \cite{hannun2019tds} trained using CTC loss \cite{graves2006connectionist} on the LibriSpeech dataset using the orthographic transcription. The trained model was used to perform inference (greedy frame-by-frame decoding) with the wav2letter++\cite{wav2letter++} audio analysis pipeline on the unlabeled audio by mapping all of the letters to \textsc{SPEECH} and the silence symbol to \textsc{NONSPEECH}. The posterior \textsc{SPEECH} probability is added as meta data on the JSON of each file.

The TDS model used for VAD has 100 million parameters and consists of clusters of 2, 3, 4, and 5 TDS blocks separated by 2D convolutions. The duration of audio contained in each label is dependent on the stride of the underlying acoustic model; the model used has a stride of 8. The model is trained with word-pieces using the recipe outlined in \cite{hannun2019tds}.

\subsubsection{SNR}\label{sec:supsnr}
The Signal-to-noise (SNR) ratio is calculated using the VAD labels predicted above. For each 80ms frame the VAD will return a posterior of whether the frame is speech or not. We decided to take $<$ 0.8 as the speech threshold, and $>$ 0.995 as noise threshold
. If a speech frame is detected, we also automatically include 2 subsequent frames, to compensate for the spiky predictions from the VAD model. Any other frames that does not belong to either bucket are ignored since we are not confident whether they are speech or noise. Finally we compute the SNR ratio by using its definition $\mathrm{SNR_{dB}} = 10 \log_{10} \left ( \frac{P_\mathrm{signal}}{P_\mathrm{noise}} \right )$

\begin{figure}[h]
  \centering
  \centerline{\includegraphics[width=6.0cm]{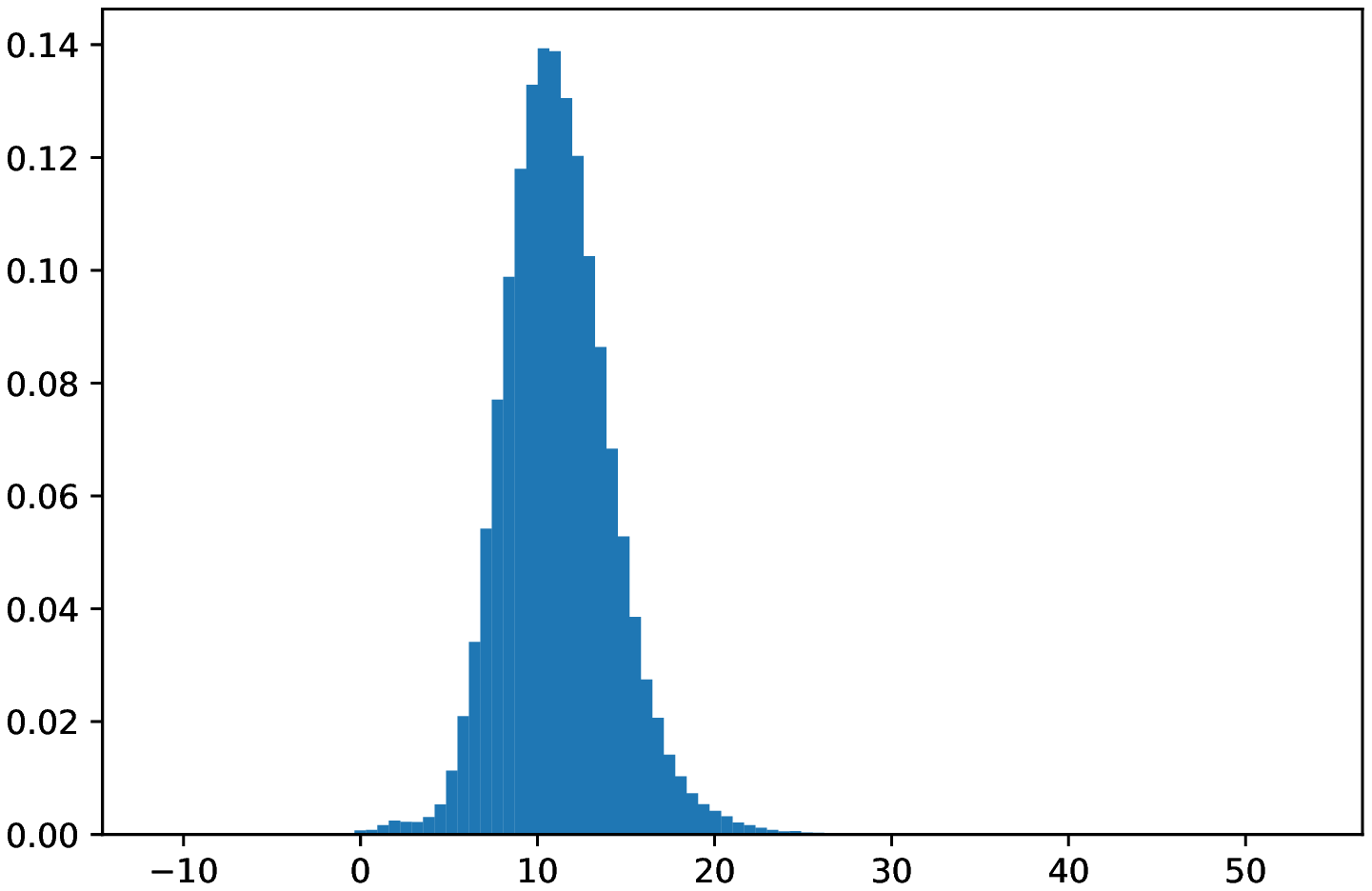}}
  \vspace{-1.3mm}
\caption{\textbf{Librivox SNR histogram}}
\label{fig:librivox_snr}
\end{figure}

\subsubsection{Perplexity}\label{sec:perplex}
Perplexity was obtained by performing beam search decoding of the trained TSD model defined above supplemented by a 4gram word Language Model trained on LibriSpeech LM. It was computed as the mean of the log probability of the posterior on each file. 

\subsubsection{JSON files and splits}\label{sec:json}

To construct the JSON, we duplicated the metadata of the original book JSON file from LibriVox into each of the files associated for a given book (including unique book ID and speaker ID, and we added tags for SNR, perplexity and our own macro-genre tags, by folding the existing ones into 7 categories: "Literature", "Science, Craft \& Essay", "Ancient", "Religion", "Poetry", "Theater", and "Undefined". We also added VAD information as a list of onsets and offsets of voice activity.

The files were splitted into cuts of different sizes by trying to maintain the same distribution of macro-genres in the three cuts. 

\subsection{The ABX metric}

Given a frame-wise distance metric, we want to check that features coding for the same phonemes have a more similar representation than features coding for different phonemes.
To quantify this property, we use a minimal pair ABX task as defined in \cite{Schatz2013ABX}; given a set of sounds $S(\boldsymbol{x})$ from a category $\boldsymbol{x}$ and a set of sounds $S(\boldsymbol{y})$ from a category $\boldsymbol{y}$ we compute:

\[ \boldsymbol{\theta}(\boldsymbol{x},\boldsymbol{y}) = \frac{1}{nm(m-1)}\sum_{a \in S(\boldsymbol{x})} \sum_{b \in S(\boldsymbol{y})} \sum_{c \in S(\boldsymbol{x}) \backslash \{ a \}} \hat{\theta}(a,b,c)\]

With:
\[ \hat{\theta}(a,b,c) = \mathbbm{1}_{d(a,c) < d(a,b)} + \frac{1}{2}  \mathbbm{1}_{d(a,c) = d(a,b)}\]
\[ m, n = |S(\boldsymbol{x})|, |S(\boldsymbol{y})| \]

Here, $\boldsymbol{\theta}(\boldsymbol{x},\boldsymbol{y})$ is the probability of a sample from category $\boldsymbol{x}$ to be closer to another element from $x$ than to one from$\boldsymbol{y}$.
To compute ABX we average the error $1-\boldsymbol{\theta}$ over all categories.

\section{Supplementary baseline models}\label{sec:supbaseline}

\subsection{CPC model and training}

To train a feature model in an unsupervised fashion, we used the method implemented by Riviere and al. in \cite{universal_speech_features}, inspired by the Contrastive Predictive Coding algorithm (CPC) presented in \cite{vandenoord2018}.
We will briefly introduce the algorithm in this section, though we refer the reader to the original papers for more details.

\subsubsection{Contrastive Predictive Coding}

CPC relies on forward modeling: given an input sequence of features, we try to predict the $k$ future representations of the sequence. 
The network must discriminate each future ground truth feature from negative examples randomly sampled in the batch.

More precisely, the model goes like this:
\begin{enumerate}
    \item The raw waveform $w$ goes through a convolutional network $g_c$, resulting in a feature sequence $(\mathbf{x}_t)_{t \in 1... T}$.
    \item Then we form the current phoneme representation $\mathbf{z}_t$ by applying a recurrent network $g_{ar}$ to $\mathbf{x}_t$.
    \item Finally, we predict $(\mathbf{x}_{t_0 +1},..., \mathbf{x}_{t_0 +k})$ from $(\mathbf{z}_t)_{t \leq t_0}$ using a prediction network $g_p$.
\end{enumerate}

When using theses feature for another task we always consider $\mathbf{z}_t$, the output of the recurrent layer.

\subsubsection{Architecture Details}

For $g_c$, we use five convolutional layers with strides [5, 4, 2, 2, 2], filter-sizes [10, 8, 4, 4, 4] and 256 hidden units with ReLU activations. 
Besides, the features are normalized channel-wise between each convolution.
In the end, this network has a downsampling factor of 160, meaning that on a $16$kH audio input each feature will encode $10$ms of data.
Furthermore, $g_{ar}$ is a one-layer LSTM with also a $256$ dimensional hidden state.
Finally, the predictor $g_p$ is a one-layer transformer.

\subsubsection{Training Details}

We considered input sequences of $1280$ms gathered in batches of $32$ sequences per GPU with a total of $128$ GPUs.
Our training took approximatly two days on NVIDIA Tesla V100-SXM2-16GB.
Besides, in a given batch, all sequences were sampled within the same speaker.

\subsection{TDS model and training}
Given the limited amount of supervised training data, we select to use a smaller TDS model \cite{hannun2019tds} with 20 million parameters. The model has a stride 2 in the first convolution, and three groups of TDS blocks with channels (10, 14, 18) and (2, 2, 3) blocks in each group. While on the whole 60k hours training data, we use the original architecture introduced in \cite{hannun2019tds} with 37 million parameters. The only difference is that we reduce the overall stride from 8 to 2. We use dropout to prevent over-fitting, and its value is set to 0.4 and 0.1 in the 20M and 37M models.

In terms of model optimization, we use plain SGD with momentum. The initial learning rate and momentum are set to 0.1 and 0.5 respectively. In the supervised setting, the models are trained for 1500 epochs on 8 GPUs in total with learning rate halved after each 200 epochs and total batch size 64 and 16 for character- and phone- based system. In the semi-supervised scenario, the models are trained for 150 epochs on 32 GPUs in total with learning rate halved after every 30 epochs and total batch size 256.

The beam-search decoding parameters are tuned on only dev-other for the 20M TDS models to generate pseudo-labels, while they are tuned independently for the final 37M model. The same official LibriSpeech 4-gram LM is used in both decoding procedures. The decoding beam size is 1000 in all the experiments. 

\section{Supplementary Results}\label{sec:supres}
\subsection{Pseudo-labels experiment}
\begin{table}[h]
\begin{tabular}{l c @{\hspace{0.8\tabcolsep}} c @{\hspace{0.8\tabcolsep}} c @{\hspace{0.8\tabcolsep}} c }
\toprule
                          & dev-      & dev-    & test-     & test-\\
System                    & clean     & other   & clean     & other\\
\midrule 
MFSC TDS + train-1h        &    44.4   &    57.7 &     55.6      &      65.9    \\
~~ + 60k pseudo-label      &    57.6   &    68.1 &     59.5      &      72.3    \\
MFSC TDS + train-10h       &    22.5   &    40.2 &     22.2      &      41.3    \\
~~ + 60k pseudo-label      &\bf 18.7   &\bf 36.0 & \bf 18.7      &\bf   38.6    \\
\midrule
\midrule
MFSC TDS + train-1h       &    44.3   &    53.9 &     46.9   &  55.4    \\
~~  + 60k pseudo-label    &    43.4   &    52.6 &     43.2   &  53.9    \\
MFSC TDS + train-10h      &    20.7   &    36.4 &     21.8   &  38.0    \\
~~ + 60k pseudo-label     &\bf 15.0   &\bf 31.1 & \bf 14.7   &\bf   32.2  \\
\bottomrule
\end{tabular}
\caption{\textbf{PER/CER of acoustic models trained in with pseudo-labels.}  \textit{Top}: small phone-based TDS \cite{hannun2019tds} models with limited labels using wav2letter++\cite{wav2letter++}, generating pseudolabels on the 60K dataset with an in-domain LM, retraining a larger TDS acoustic model (PER). \textit{Bottom}: the same trained on characters (CER).} 
\label{tab:perdistant}
\end{table}

Table \ref{tab:perdistant} provides PER/CER in the distant supervision setting with models trained on pseudo-labels. The TDS models above and generated pseudo-labels are trained and generated with the exactly the same procedure introduced in Section \ref{sec:pagestyle}. Note that the PER/CER results above are not comparable to the semi-supervised ones in Table \ref{tab:semi} as pseudo-labels here are generated with the official LibriSpeech LM, whose training set is a super-set of the transcriptions in the supervised training set.  

\end{document}